\documentclass[letterpaper, 10 pt, conference]{ieeeconf}  
\IEEEoverridecommandlockouts

\usepackage[top=57pt,bottom=43pt,left=48pt,right=48pt]{geometry}
\usepackage{cite}
\usepackage{amsmath,amssymb,amsfonts}
\usepackage{algorithm}
\usepackage{booktabs} 
\usepackage{algpseudocode}
\usepackage{graphicx}
\usepackage{textcomp}
\usepackage{xcolor}
\usepackage{subfiles}
\usepackage{mathtools}
\usepackage{subcaption}
\usepackage{graphicx}
\usepackage{svg}
\usepackage{url}
\usepackage{hyperref}
\usepackage{comment}
\newcommand\norm[1]{\lVert#1\rVert}

\DeclareMathOperator*{\argmin}{arg\,min}

\def\BibTeX{{\rm B\kern-.05em{\sc i\kern-.025em b}\kern-.08em
    T\kern-.1667em\lower.7ex\hbox{E}\kern-.125emX}}

\begin{document}
\title{\LARGE \bf Range-based Multi-Robot Integrity Monitoring Against Cyberattacks and Faults: An Anchor-Free Approach}
\author{Vishnu Vijay, Kartik A. Pant, Minhyun Cho, Yifan Guo, James M. Goppert, and Inseok Hwang
\thanks{\textcolor{blue}{This work has been submitted to the IEEE for possible publication. Copyright may be transferred without notice, after which this version may no longer be accessible.}}
\thanks{This research is funded by the Secure Systems Research Center (SSRC) at the Technology Innovation Institute (TII), UAE. The authors are grateful to Dr. Shreekant (Ticky) Thakkar and his team members at the SSRC for their valuable comments and support.}
\thanks{The authors are with the School of Aeronautics and Astronautics, Purdue University,
West Lafayette, IN 47906. Email: ({\tt\small vvijay}, {\tt\small kpant}, {\tt\small cho515}, {\tt\small guo781}, {\tt\small jgoppert}, {\tt\small ihwang}){\tt\small @purdue.edu}}%
}

\maketitle
\begin{abstract}
Coordination of multi-robot systems (MRSs) relies on efficient sensing and reliable communication among the robots. However, the sensors and communication channels of these robots are often vulnerable to cyberattacks and faults, which can disrupt their individual behavior and the overall objective of the MRS. In this work, we present a multi-robot integrity monitoring framework that utilizes inter-robot range measurements to (i) detect the presence of cyberattacks or faults affecting the MRS, (ii) identify the affected robot(s), and (iii) reconstruct the resulting localization error of these robot(s). The proposed iterative algorithm leverages sequential convex programming and alternating direction of multipliers method to enable real-time and distributed implementation. Our approach is validated using numerical simulations and demonstrated using PX4-SiTL in Gazebo on an MRS, where certain agents deviate from their desired position due to a GNSS spoofing attack. Furthermore, we demonstrate the scalability and interoperability of our algorithm through mixed-reality experiments by forming a heterogeneous MRS comprising real Crazyflie UAVs and virtual PX4-SiTL UAVs working in tandem.   

\end{abstract}

\section{Introduction}
\label{sec:intro}
Recent advances in sensing, networking, planning, and control, as well as the development of high-performance computational hardware, have enabled the deployment of robotic systems for real-world applications. Within robotics, a multi-robot system (MRS) refers to the coordination and teaming of more than one robot to accomplish complex tasks autonomously without any human supervision. Such collaboration allows the system to solve complicated problems that could not have been possible by a single robot; it has been widely used in warehouse logistics \cite{claes2017decentralised}, vehicle platooning \cite{axelsson2016safety}, connected vehicle-to-vehicle operations \cite{li2017platoon,jia2015survey}, surveillance \cite{van2008non}, and disaster relief \cite{ghassemi2022multi}. 

However, the reliance on autonomous MRS often raises concerns about safety, security, and reliability in real-world settings. Furthermore, these systems present security challenges as malicious actors (i.e., hackers) can alter sensor measurements and/or jam the communication channels to disrupt their operations. The presence of an attacked or faulty robot in the network could potentially manipulate the entire MRS, leading to catastrophic failures and harming critical infrastructure, such as UAVs crashing into buildings, etc. To maintain the integrity of the system, it is essential to ensure (i) safety (i.e., avoiding collision or reaching an undesired state) and (ii) liveliness (i.e., the ability to complete the task without any disruption). Thus, detecting and identifying rogue or faulty robots and isolating them becomes essential for safe and efficient multi-robot operations.
\begin{figure}[t!] 
\centering
  \includegraphics[scale=0.5, keepaspectratio]{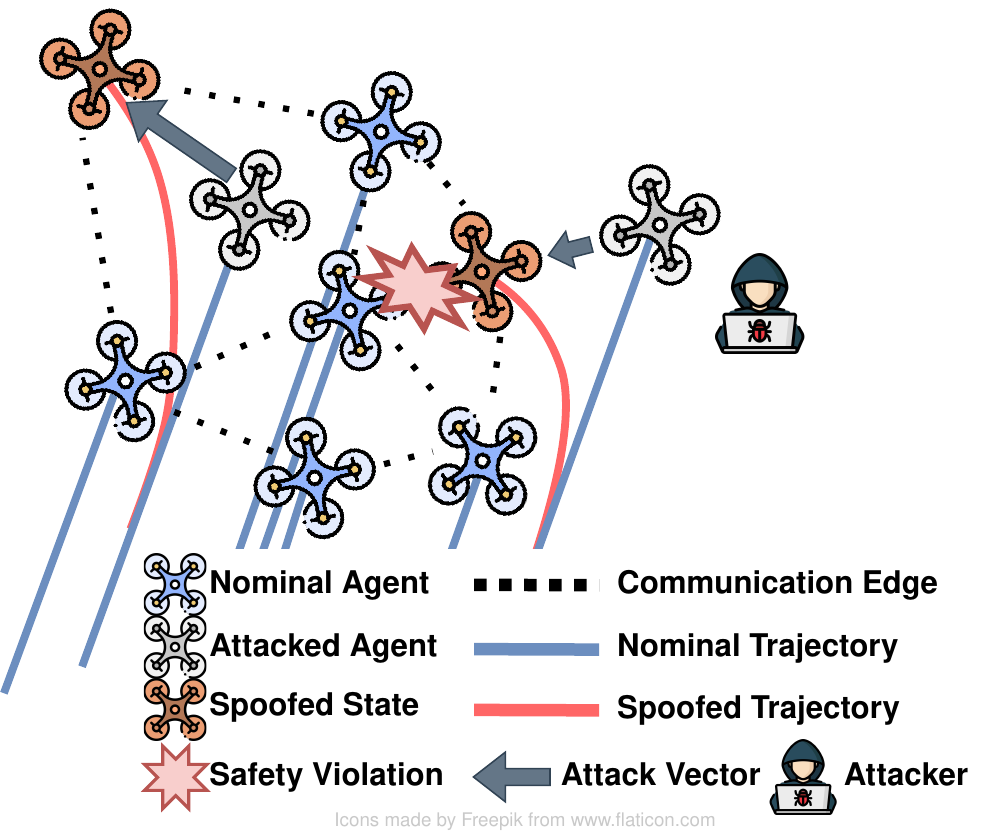}
  \caption{Illustration of spoofing attacks in a multi-robot system}
  \label{part1:description}
\end{figure}

Unlike a single-robot system, in an MRS, there exists an inherent redundancy of information in the form of locally sensed or communicated information from neighbors. This external information can be leveraged to enhance the robustness of the MRS against cyberattacks and system faults. One such type of information is inter-robot measurements between agents of the MRS. 
However, combining these external inputs from other robots (healthy or otherwise) in a real-time distributed fashion is challenging. One main challenge is how we trust which robots are relaying correct information. 
Existing works on robust multi-robot anomaly detection have been focused on an anchor-based approach \cite{zhao2019bearing}, where some of the robots are considered anchors whose positions are assumed correct, which may not always be possible. In this work, we consider an anchor-free approach that utilizes information from each robot to identify and localize errors in the system. We propose a distributed runtime integrity monitor for an MRS that leverages inter-robot range measurements to indiscriminately detect and identify rogue or faulty robots. We design an iterative algorithm that solves the nonlinear integrity monitoring optimization problem using sequential convex programming (SCP). This SCP problem can be solved efficiently in a distributed manner using the alternating direction of multiplier method (ADMM) \cite{boyd2011distributed}.    

We introduce two different metrics to assess the integrity of an MRS: first, \textit{overall system integrity}, which evaluates the integrity of the overall MRS to inform system error detection, and second, \textit{robot integrity}, which assesses the integrity of each robot to inform robot error identification. Through extensive numerical simulations, we demonstrate the robustness of our proposed algorithm, which is subject to errors in the state estimation and noise in the inter-robot range measurements. We also devise a threshold-based mechanism to bypass the warm start method of ADMM and reset internal parameters of the optimization problem, which specifically addresses the time-varying network topology of an MRS. This enables our algorithm to more effectively adapt to changes in its network topology. For example, in a multi-robot inspection and surveillance mission, if a certain number of robots are identified as attacked or faulty and the system's network topology is reconfigured, our proposed approach will adjust the fault monitor by resetting the algorithm's internal parameters and bypassing warm start.

In this paper, our main contributions are as follows,
\begin{itemize}
    \item We propose a novel anchor-free multi-robot integrity monitoring framework using inter-robot range measurements to detect cyberattacks or system faults that could go unnoticed by traditional single-robot fault detectors.
    \item We validate the performance and robustness of the proposed algorithm through extensive numerical simulations with noise, as well as a time-varying network topology resulting from a cyberattack or fault.
    \item We create a Mixed Reality inter-robot range sensor emulation by leveraging the physics engine of Gazebo.   
    \item Finally, we demonstrate the effectiveness of the proposed algorithm through Mixed Reality experiments on a heterogeneous MRS comprised of real Crazyflie UAVs and virtual PX4-SiTL.   
\end{itemize}

The rest of the paper is organized as follows. Section \ref{sec:related_work} summarizes related works in multi-agent fault detection and isolation. Section \ref{sec:background} reviews the background concepts from graph theory, control theory, and optimization utilized to set up the multi-robot integrity monitoring problem. In Section \ref{sec:sys_arch}, we present the details of the algorithm, including a sensitivity analysis of the robustness of the proposed approach subject to noise and a time-varying network topology. Section \ref{sec:sim} presents the numerical simulation results and elucidates the effect of the algorithm's parameters on the convergence of the algorithm. Section \ref{sec:exp} describes the experimental validation of the proposed algorithm using Mixed-Reality-in-the-loop (MRiTL) experiments comprised of real Crazyflie UAVs and virtual PX4-SiTL. Finally, Section \ref{sec:conclusion} concludes the paper and presents future directions.

\section{Related Works}
\label{sec:related_work}
Multi-agent fault detection and identification is a topic that has fostered exciting research in several fields. In control theory, significant efforts have been made to theoretically analyze multi-agent fault detection and isolation algorithms using approaches such as $H_\infty$ performance indices \cite{chadli2017distributed, gallehdari2017h}, residual testing \cite{guo2012distributed}, $l_1$ norm minimization \cite{hashimoto2019distributed}, and interval observers \cite{zhang2018distributed}. These works extend the scope of fault detection from single-agent to multi-agent systems; however, they are limited to linear measurement models and do not accommodate nonlinear measurement models. In our earlier work \cite{khan2023collaborative, khan2023recovery}, we proposed a rigidity theory approach that can be applied to various nonlinear measurement models (e.g., range, bearing, etc.). We focused on providing a theoretical foundation for anchor-free multi-agent fault detection and isolation while considering an idealistic noise-free setting and static network topology. We extend the work in \cite{khan2023collaborative} to enable its use for practical applications, with a focus on robustness to noise and time-varying network topologies.

Active investigation on anomaly detection is also proceeding in the machine learning community, where researchers blur the boundary between single-agent and multi-agent problems and detect anomalies via only signal patterns.
Three major categories of approaches are prediction \cite{tan2020lstm, hao2021hybrid}, reconstruction \cite{bhatia2019unsupervised}, and classification \cite{lin2019dynamic,goh2017anomaly,shen2020timeseries}. These machine learning-based methods can usually outperform control theory-based methods on benchmark datasets. However, these approaches do not provide any theoretical interpretation or guarantees, making them unsuitable for safety-critical applications. These approaches also don't consider the communication constraints of real systems, which can significantly affect the integrity monitoring performance. Our proposed method, however, achieves real-time anomaly detection in a distributed manner with convergence guarantees. 

In the robotics community, researchers combine ideas from both control and machine learning areas to focus on real-world multi-robot systems, where the system dynamics, communication ability, power conditions, etc., are further specified \cite{tarapore2019fault, suarez2016cooperative, sindhwani2020unsupervised}. Such specifications can facilitate the validation of their algorithms on real-world robotic systems, but may also limit the scope of their work to a specific MRS. For example, Suarez et al. \cite{suarez2016cooperative} proposed a voting-based fault detection algorithm for a multi-UAV system where each UAV raises a flag on its direct neighbors when the discrepancy between independent state estimators surpasses a threshold. However, such an approach requires at least two perfectly healthy UAVs that can play as anchors (attack-free) equipped with independent estimators, 
which might not be practical as the multiple robots in a swarm can be attacked simultaneously. Lee et al. \cite{lee2022data} proposed a data-driven approach to select the most effective fault detection metric. However, their method can only select from a pre-defined metric set, which may differ substantially for each system and limits its applicability to a general multi-robot system. On the contrary, our proposed approach applies to a wide range of inter-robot measurements accessible for most multi-robot systems. Other representative works in robotics for multi-robot fault detection and isolation include \cite{arrichiello2014distributed, zhang2018distributed, tarapore2019fault, kutzer2008toward,sindhwani2020unsupervised}, which study fault detection in robotic systems but don't consider cyberattacks, which are carefully designed to evade statistical fault detection methods.

\section{Background}
\label{sec:background}

\subsection{Notation}

The set of real numbers and integers are denoted as $\mathbb{R}$, $\mathbb{Z}$, and the superscript $+$ stands for the non-negativeness of the set. Then, the vector $\mathbf{v}[i] \in \mathbb{R}^{\ell}$ refers to the $i^\text{th}$ block of a block vector $\mathbf{v} \in \mathbb{R}^{k \ell}$. Then, the $l_{q}$ norm of the block vector is denoted as follows:
\begin{align*}
    \norm{\mathbf{v}}_{2,q} = \begin{cases}
        \sum_{i=1}^{k} \mathbb{I} \left(\norm{\mathbf{v}[i]}_2 > 0 \right) & \text{when} \quad q=0 \\
        \left( \sum_{i=1}^{k} \norm{\mathbf{v}[i]}_2^q \right)^{1/q} & \text{when} \quad 0<q<\infty \\
        \max_{1 \leq i \leq k} \left( \norm{\mathbf{v}[i]}_2 \right) & \text{when} \quad q = \infty,
    \end{cases}
\end{align*}
where $\mathbb{I}(\cdot)$ is an indicator function that yields $1$ when the entity is positive and $0$ otherwise. Therefore, the vector's $l_{0}$ norm represents the block vector's non-zero blocks, i.e., block sparsity. For a given index set $\mathcal{S}$, we denote the cardinality of the set as $|\mathcal{S}|$; in other words, $\mathcal{S}=\left\{1,2,\cdots,|\mathcal{S}|\right\}$. 

\subsection{Sensing and Communication Network}
We consider a multi-robot system described by a graph $\mathcal{G}$, composed of a set of vertices $\mathcal{V}$ and a set of edges $\mathcal{E}$. With the graph vertices representing the agents of the multi-robot system, the edges represent the measurements that can be computed between the agents. Two robots are considered neighbors if an edge connects them, and the neighbors of robot $i \in \mathcal{V}$ are included in set $\mathcal{N}_i$. Using a similar notation, the set of edges connecting robot $i \in \mathcal{V}$ is denoted by $\mathcal{E}_i$. This structure can be generalized to a hypergraph with hyperedges to allow for the use of inter-robot measurement models that connect three or more robot, such as time difference-of-arrival.
An important assumption made in the MRS communication network is that neighboring robot can communicate synchronously with each other. 

\subsection{Dynamics, Consensus Protocol and State Estimation}
Defining the state of the $i^\text{th}$ agent of the MRS to be $\mathbf{p}[i] \in \mathbb{R}^{n_i}$, the collective state of the system is represented by the block vector $\mathbf{p} \in \mathbb{R}^{n}$ 
\begin{equation}
    \mathbf{p} = 
    \begin{bmatrix}
        \mathbf{p}[1]^\top & \mathbf{p}[2]^\top & ... & \mathbf{p}[|\mathcal{V}|]^\top
    \end{bmatrix}^\top
\end{equation}

Each robot in the MRS, corresponding to the vertices $\mathcal{V}$, is governed by dynamics and a consensus control law with a state estimator.  The $i^\text{th}$ agent dynamics is given as the following linear discrete-time system:
\begin{align} \label{part3: agent dynamics}
    \begin{cases}
        \mathbf{p}[i](k+1) = A_{i}\mathbf{p}[i](k) + B_{i}\mathbf{u}[i](k)+E_{i}\mathbf{w}[i](k)\\
        \mathbf{q}[i](k) = C_{i}\mathbf{p}[i](k) + F_{i}\mathbf{v}[i](k)+\Gamma_{i}\mathbf{f}[i](k),
    \end{cases}
\end{align}
where  $\mathbf{p}[i](k) \in \mathbb{R}^{n_{i}}$ is the state of agent $i$, $\mathbf{u}[i](k) \in \mathbb{R}^{u_{i}}$ is the control input, $\mathbf{q}[i](k) \in \mathbb{R}^{q_{i}}$ is the measurement output, $\mathbf{w}[i](k) \in \mathbb{R}^{w_{i}}$ and $\mathbf{v}[i](k) \in \mathbb{R}^{v_{i}}$ are the process and measurement noises satisfying the norm bounded condition:
\begin{align} \label{part3: bounded noises}
    \begin{split}
        &\norm{\mathbf{w}[i](k)}_2^2 = \mathbf{w}[i](k)^\top \mathbf{w}[i](k) \leq W_{i} \\
        &\norm{\mathbf{v}[i](k)}_2^2 = \mathbf{v}[i](k)^\top \mathbf{v}[i](k) \leq V_{i}
    \end{split}
\end{align}
with $W_{i},V_{i} \in \mathbb{R}^+$ and $k \in \mathbb{Z}^+$. The matrices $A_{i} \in \mathbb{R}^{n_{i} \times n_{i}}$, $B_{i} \in \mathbb{R}^{n_{i} \times u_{i}}$ and $C_{i} \in \mathbb{R}^{q_{i} \times n_{i}}$ denote the system matrix, the input matrix, and the output matrix, respectively. The matrices $E_{i} \in \mathbb{R}^{n_{i} \times w_{i}}$ and $F_{i} \in \mathbb{R}^{q_{i} \times v_{i}}$ are the perturbation matrices of the process and measurement noises, respectively. The attack or fault input is denoted as  $\mathbf{f}[i](k) \in \mathbb{R}^{f_{i}}$ with the associated attack matrix $\Gamma_{i} \in \mathbb{R}^{q_{i} \times f_{i}}$. The system pairs $\left(A_i, B_i \right)$ and $\left(A_i, C_i \right)$ are assumed to be stabilizable and detectable. Each robot estimates their state using their own measurements $\mathbf{q}[i](k)$ and the following state estimator:
\begin{align} \label{part3: state estimator}
    \begin{cases} 
        \hat{\mathbf{p}}[i](k+1) =  A\mathbf{p}[i](k) + B\mathbf{u}[i](k) \\ \qquad \qquad \qquad \qquad + L_{o,i}\left(\mathbf{q}[i](k) - \hat{\mathbf{q}}[i](k) \right) \\
        \hat{\mathbf{q}}[i](k) = C \hat{\mathbf{p}}[i](k),
    \end{cases}
\end{align}
where $L_{o,i} \in \mathbb{R}^{n_{i} \times q_{i}}$ is the estimator gain, $\hat{\mathbf{p}}[i](k) \in \mathbb{R}^{n_{i}}$ is the estimated state of agent and $\hat{\mathbf{q}}[i](k) \in \mathbb{R}^{q_{i}}$ is its estimated output. Then, the consensus control protocol that governs the behavior of the MRS is given as:
\begin{align} \label{Part2: consensus law}
     & \mathbf{u}[i](k) = K_{c,i} \Bigl(\sum_{j \in \mathcal{N}_i} a_{ij} (\hat{\mathbf{p}}[i](k)-\hat{\mathbf{p}}[j](k))\Bigr) + \mathbf{u}_{r}(k),
\end{align}
where $K_{c,i} \in \mathbb{R}^{n_{i} \times u_{i}}$ is the consensus gain, $\mathbf{u}_{r}(k)$ is the reference trajectory tracking input and $a_{ij}$ is the element of the adjacency matrix $\mathcal{A}$. Since we assume an undirected communication network in the problem, $a_{ij} = 1$ if the agent $j$ is a neighbor of $i$, i.e., $j \in \mathcal{N}_{i}$, $a_{ij} = 0$ otherwise. For simplicity, the time index $k$ will be dropped in the subsequent sections. The gain for consensus protocol and state estimator can be designed to satisfy $H_\infty$ performance criterion, which results in the optimal estimation with a bounded error norm, i.e., $\norm{\mathbf{p}-\hat{\mathbf{p}}}_{2}=\norm{\pmb{\nu}}_{2} \leq \nu_\text{max}$ when the system is without attack or fault, i.e., $\mathbf{f}[i] = 0$.  The consensus gain and the estimator gain can be implemented by solving LMI \cite{hwang2023lmi,cho2024risk}. We omit the detailed implementation since it is not in the scope of the paper.

\subsection{Attack or Fault Model}

We define the error vector $\mathbf{x} \in \mathbb{R}^n$ as the discrepancy between the true and estimated states resulting from a cyberattack or fault. This gives $\mathbf{x} = \mathbf{p} + \pmb{\nu} - \hat{\mathbf{p}}$ for a system with localization error. Defining $\mathcal{D}$ to be the set of affected robots, we see $| \mathcal{D} | = \norm{\mathbf{x}}_{2,0}$. We further assume the error vector $\mathbf{x}$ is sparse, or $| \mathcal{D} | \ll | \mathcal{V} |$. This is a reasonable assumption as, in practice, attackers will have a limited budget and faults do not occur frequently. Finally, we denote the desired estimated error $\hat{\mathbf{x}} = \mathbf{p} - \hat{\mathbf{p}}$.  

\subsection{Inter-Robot Measurements}

The measurement model $\mathbf{\Phi} \in \mathbb{R}^m$ is a block vector with each subvector corresponding to an edge. The subvectors are denoted by $\mathbf{\Phi}^{(l)} \in \mathbb{R}^{m_l}$ for $l \in \{0, 1, ..., |\mathcal{E}|\}$. The true inter-robot measurement $\mathbf{y} \in \mathbb{R}^m$ can be found according to $\mathbf{y} = \mathbf{\Phi}(\mathbf{p})$. There are various types of inter-robot measurements for MRS implementation; for the purposes of this work, we use range measurements due to the ease and cost of implementation. For example, Ultra-Wide Band (UWB) sensors require low processing for range computations. Other types of inter-robot measurements may require more complex sensors that would necessitate more processing power.

Each robot measures the range to its neighbors using noisy sensors. Letting the block vector $\pmb{\omega} \in \mathbb{R}^m$ represent the range sensor noise, we define a noisy inter-robot range measurement $\hat{\mathbf{y}} = \mathbf{y} + \pmb{\omega}$. We assume the noise is norm bounded such that $\norm{\pmb{\omega}}_2 \leq \omega_\text{max}$. When $\mathbf{p} \neq \hat{\mathbf{p}}$ in a noisy, attacked, and/or faulty system, it can be seen that $\mathbf{y} \neq \mathbf{\Phi}(\hat{\mathbf{p}})$. We recognize that including error and noise will result in
\begin{equation}
\hat{\mathbf{y}} = \mathbf{\Phi}(\hat{\mathbf{p}} + \hat{\mathbf{x}}) + \pmb{\omega}.    
\end{equation}
Thus, the inter-agent measurement model $\mathbf{\Phi}$ can be leveraged to solve for $\hat{\mathbf{x}}$. We assume no knowledge of $\hat{\mathbf{x}}$ and solve for the error term in a process known as \textit{error reconstruction}.

\section{Anchor-Free Fault Detection and Reconstruction}
\label{sec:sys_arch}

\begin{figure}[t] 
\centering
  \includegraphics[width=0.45\textwidth]{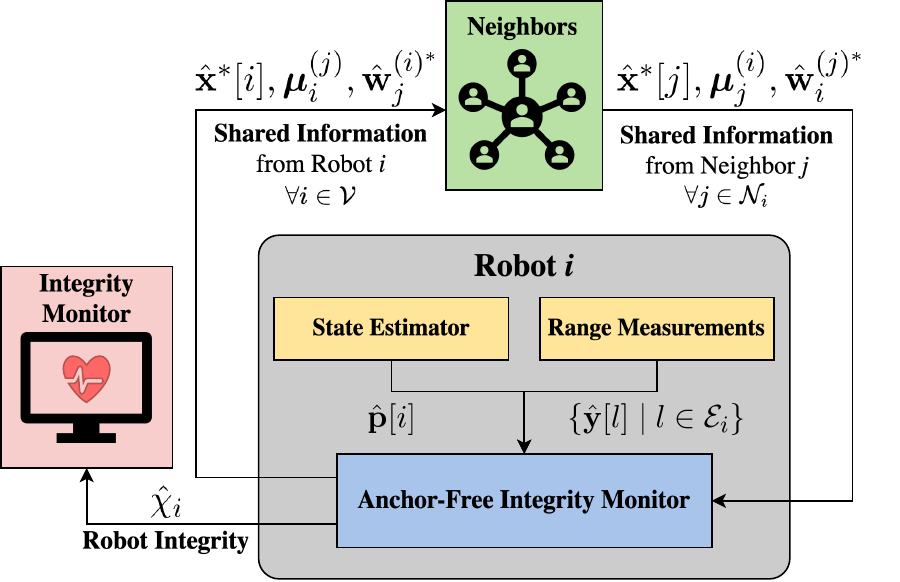}
  \caption{Overall System Architecture}
  \label{fig:sys_arch}
\end{figure}
In this section, we describe the steps required to estimate the true error vector $\mathbf{x}$ using robot state estimates and noisy inter-robot range measurements. To do so, we first present the optimization problem, which is relaxed using the SCP and ADMM techniques, and then discuss parameter tuning required for real-time and distributed implementation. We define a threshold so only significant estimated error vectors $\hat{\mathbf{x}}$ are considered, while others are attributed to noise or transient behavior. We then address challenges associated with implementing such an algorithm on a real-world system, including noise and time-varying network topology. The overall architecture of our proposed MRS integrity monitor is described in Fig. \ref{fig:sys_arch}.

\subsection{Error Reconstruction}
We adopt the same problem formulation as presented in \cite{khan2023collaborative} and seek to find the sparsest error vector (i.e., block error vector with least nonzero subvectors) that explains the inconsistency in our measurement model $\mathbf{\Phi}(\hat{\mathbf{p}}) \neq \hat{\mathbf{y}}$. Under noise-free setting \cite{khan2023collaborative}, this problem can be posed as 
\begin{equation}
    \label{eq:P1}
    \begin{split}
        \min_{\hat{\mathbf{x}} \in \mathbb{R}^n} & \hspace{20pt} \norm{\hat{\mathbf{x}}}_{2,0} \\
        \text{s. t.} & \hspace{20pt} \mathbf{\Phi} (\hat{\mathbf{p}} + \hat{\mathbf{x}} ) = \hat{\mathbf{y}}
    \end{split}
\end{equation}
As the objective function and the constraints are nonlinear, the optimization problem \eqref{eq:P1} cannot be solved efficiently in polynomial time. Also, the coupling of information among robots in the constraints doesn't allow \eqref{eq:P1} to be solved in a distributed manner. Thus, a relaxed, linearized problem is formulated by applying the SCP and ADMM algorithms.

We first apply SCP, which solves successive local convex optimization problems, by re-linearizing the constraints at each iteration. Let $\Bar{\mathbf{x}}$ be the minimized error vector from the convexified sub-problem. The $\Bar{\mathbf{x}}$ from successive iterations are summed to reconstruct the error vector. Note that $\hat{\mathbf{x}}$ is the local error vector being minimized. By replacing nonconvex $\norm{\cdot}_{2,0}$ with convex $\norm{\cdot}_{2,1}$, we introduce the convex objective function $\norm{\hat{\mathbf{x}} + \Bar{\mathbf{x}}}_{2,1}$. 
Additionally, the constraint can be rewritten using a first-order Taylor approximation of the measurement model:
we let $\mathbf{z} = \hat{\mathbf{y}} - \mathbf{\Phi} (\hat{\mathbf{p}} + \Bar{\mathbf{x}})$ and $\mathbf{R} = \mathbf{J}_\mathbf{\Phi} ( \hat{\mathbf{p}} + \Bar{\mathbf{x}} ) $ to form a linearized constraint $\mathbf{R} \hat{\mathbf{x}} = \mathbf{z}$.

This can be further transformed using ADMM, which decouples the optimization problem by breaking it down into smaller parts and is well suited for distributed convex optimization \cite{boyd2011distributed}. At each robot $i \in \mathcal{V}$, we introduce a new set of primal variables $\hat{\mathbf{w}}_i^{(j)}$ that act as copies of the state of robot $i$ for each neighbor $j \in \mathcal{N}_i$. Consistency constraints are defined to ensure $\{ \hat{\mathbf{x}}_i = \hat{\mathbf{w}}_i^{(j)} \mid j \in \mathcal{N}_i \}$. The objective function is augmented with quadratic constraint penalty terms to define a new optimization problem:
\begin{equation}
    \label{eq:P3}
    \begin{split}
        \min_{ \hat{\mathbf{x}}, \hat{\mathbf{w}} } \hspace{15pt} & \norm{ \hat{\mathbf{x}} + \Bar{\mathbf{x}}}_{2,1} + \frac{\rho}{2}\sum_{i \in \mathcal{V}} \sum_{l \in \mathcal{E}_i} \norm{\mathbf{c}_i^{(l)} (\hat{\mathbf{x}}, \hat{\mathbf{w}})}_2^2 \\
        & + \frac{\rho}{2} \sum_{i \in \mathcal{V}} \sum_{j \in \mathcal{N}_i} \norm{\mathbf{d}_i^{(j)} (\hat{\mathbf{x}}, \hat{\mathbf{w}})}_2^2 \\
        \text{s. t.} \hspace{15pt} & \{ \mathbf{c}_i^{(l)} (\hat{\mathbf{x}}, \hat{\mathbf{w}}) = \mathbf{0} \mid l \in \mathcal{E}_i,  i \in \mathcal{V} \} \\
        & \{ \mathbf{d}_i^{(j)} (\hat{\mathbf{x}}, \hat{\mathbf{w}}) = \mathbf{0} \mid j \in \mathcal{N}_i,  i \in \mathcal{V} \}
    \end{split}
\end{equation}
where $\rho \in \mathbb{R}^+$ is known as a penalty parameter \cite{khan2023collaborative, boyd2011distributed}, $\mathbf{c}(\cdot)$ is the discrepancy between the expected and true range measurements, and $\mathbf{d}(\cdot)$ is the discrepancy between the local and neighbor reconstructed error vectors, defined as
\begin{equation}
    \label{eq:P3_constraints}
    \begin{split}
        \mathbf{c}_i^{(l)} (\hat{\mathbf{x}}, \hat{\mathbf{w}}) &= \mathbf{R}[l, i] \hat{\mathbf{x}}[i] - \Big( \mathbf{z}[l] -\sum_{j \in \mathcal{N}_i} \mathbf{R}[l, j] \hat{\mathbf{w}}_j^{(i)}  \Big)  \\
        \mathbf{d}_i^{(j)} (\hat{\mathbf{x}}, \hat{\mathbf{w}}) &= \hat{\mathbf{x}}[i] - \hat{\mathbf{w}}_i^{(j)}.
    \end{split}
\end{equation}
The choice of $\rho$ is vital to ADMM's convergence and is discussed in Section \ref{subsec:noise_effect}. We define the dual variables $\pmb{\lambda} = \{ \pmb{\lambda}_i^{(l)} \in \mathbb{R}^{m_l} \mid l \in \mathcal{E}_i, i \in \mathcal{V} \}$ and $\pmb{\mu} = \{ \pmb{\mu}_i^{(j)} \in \mathbb{R}^{n_i} \mid j \in \mathcal{N}_i, i \in \mathcal{V} \}$ for the constraints of \eqref{eq:P3} to construct the Lagrangian of the constraint optimization problem in \eqref{eq:P3_lagrangian}. For brevity, we omit the arguments of $\mathbf{c}(\cdot)$ and $\mathbf{d}(\cdot)$:
\begin{equation}
    \label{eq:P3_lagrangian}
    \begin{split}
        L(\hat{\mathbf{x}}, \hat{\mathbf{w}}, \pmb{\lambda}, \pmb{\mu}) & = \sum_{i \in \mathcal{V}}  L_i(\hat{\mathbf{x}}, \hat{\mathbf{w}}, \pmb{\lambda}, \pmb{\mu}) \\
        L_i(\hat{\mathbf{x}}, \hat{\mathbf{w}}, \pmb{\lambda}, \pmb{\mu}) &= 
        \norm{\hat{\mathbf{x}}[i] + \Bar{\mathbf{x}}[i]}_2 \\
        & \hspace{5pt} + \sum_{l \in \mathcal{E}_i} \bigg[ \frac{\rho}{2} \norm{\mathbf{c}_i^{(l)} }_2^2 + \Big( \pmb{\lambda}_i^{(l)} \Big) ^\top \mathbf{c}_i^{(l)}  \bigg] \\
        & \hspace{5pt} + \sum_{j \in \mathcal{N}_i} \bigg[ \frac{\rho}{2} \norm{\mathbf{d}_i^{(j)} }_2^2 + \Big( \pmb{\mu}_i^{(j)} \Big) ^\top \mathbf{d}_i^{(j)} \bigg] .
    \end{split}
\end{equation}

Applying SCP and ADMM creates a nested loop structure that allows an iterative reconstruction of the error vector. Linearization of the measurement model constraint and updates to $\Bar{\mathbf{x}}$ are done in the outer loop, shown in Alg. \ref{Algorithm:Convexified_Outer_Loop}. The inner loop is composed of the ADMM framework, which handles minimization over the primal variables $\hat{\mathbf{x}}$ and $\hat{\mathbf{w}}$, as well as updates to the dual variables. This is shown in Alg. \ref{Algorithm:ADMM_Inner_Loop}. Note the constraints are linearized after $N_\text{ADMM}$ iterations of the inner loop.


\begin{algorithm}[t]
    \flushleft
    \caption{Distributed Multi-Robot FDIR for Robot $i$}
    \begin{algorithmic}[1]
        \Require (Swarm info.) $\rho$, $N_\text{ADMM}$, cold start thresholds
        \Require (Robot info.) $\{ \hat{\mathbf{y}}[l]\}_{l\in \mathcal{E}_i}$ and $\{ \hat{\mathbf{p}}[j]\}_{j\in \mathcal{N}_i}$ \\
        Letting $\mathcal{N}'_i := \mathcal{N}_i \cup \{ i \}$, initialize the following to $\mathbf{0}$: 
        $\{ \pmb{\lambda}_i^{(l)} \in \mathbb{R}^{m_l} \mid l \in \mathcal{E}_i \}$, 
        $\{ \pmb{\mu}_i^{(j)} \in \mathbb{R}^{n_i} \mid l \in \mathcal{N}_i \}$,
        $\{ \mathbf{x}^* [j] \in \mathbb{R}^{n_j} \mid j \in \mathcal{N}'_i\}$
        \While{fault monitor activated,} \\
            Linearize the constraint by computing $\{ \mathbf{z}[l] \mid l \in \mathcal{E}_i \}$ and $\{ \mathbf{R}[l,j] \mid l \in \mathcal{E}_i, j \in \mathcal{N}_i \}$ \\
            Compute $\{ \hat{\mathbf{x}}^*[j]  \}_{\forall j \in \mathcal{N}'_i} $ using Alg. \ref{Algorithm:ADMM_Inner_Loop}.\\
            Update the error vectors, $\forall j \in \mathcal{N}'_i$:
            \begin{align*}
                \Bar{\mathbf{x}}[j]^+ \leftarrow \Bar{\mathbf{x}}[j] + \hat{\mathbf{x}}^*[j]
            \end{align*} \\
            Reset dual variables if cold start flag is set
        \EndWhile
        \Ensure Reconstructed error vector $(\hat{\mathbf{x}}[i] + \Bar{\mathbf{x}}[i])$
        \Ensure Robot integrity $\hat{\chi}_i$
    \end{algorithmic}
    \label{Algorithm:Convexified_Outer_Loop}
\end{algorithm}
In many cases, it was found that retaining the values of dual variables $\pmb{\lambda}$ and $\pmb{\mu}$ between iterations of the SCP loop could promote convergence properties and is especially true for a static network topology. This technique, outlined in \cite{khan2023collaborative,boyd2011distributed}, is known as warm start. However, when this network topology is time-varying, we find that warm start struggles to promote convergence. This is further discussed in Section \ref{subsec:varying_topology}. Note that the minimization problems in Alg. \ref{Algorithm:ADMM_Inner_Loop} are computed only over the neighbors of robot $i$. Thus, the algorithm scales linearly with $| \mathcal{N}_i |$, not with $|\mathcal{V}|$. This promotes the scalability of the algorithm to large robot swarms without affecting computational complexity.
\begin{algorithm}[t]
    \flushleft
    \caption{ADMM Subroutine for Robot $i$}
    \begin{algorithmic}[1]
        \\Initialize $\hat{\mathbf{w}}_i^{(j)^*} \in \mathbb{R}^{n_i}$ and $\hat{\mathbf{w}}_j^{(i)^*} \in \mathbb{R}^{n_j}$ as $\mathbf{0}$, $\forall j \in \mathcal{N}_i$.
        \For{$N_\text{ADMM}$ iterations,} \\
            Solve first primal variable minimization: $$\hat{\mathbf{x}}^*[i] = \argmin_{\hat{\mathbf{x}}[i]} L_i(\hat{\mathbf{x}}, \hat{\mathbf{w}}^*, \pmb{\lambda}, \pmb{\mu})$$ \\

            Communicate $\hat{\mathbf{x}}^*[i]$ and $\pmb{\mu}_i^{(j)}$ to neighbors $\mathcal{N}_i$. \\
            
            Solve second primal variable minimization: $$\{ \hat{\mathbf{w}}_j^{(i)^*} \mid j \in \mathcal{N}_i \}  = \argmin_{\{ \hat{\mathbf{w}}_j^{(i)} \mid j \in \mathcal{N}_i \}} L_i(\hat{\mathbf{x}}^*, \hat{\mathbf{w}}, \pmb{\lambda}, \pmb{\mu})$$ \\

            Communicate $\hat{\mathbf{w}}_j^{(i)^*}$ to neighbors $\mathcal{N}_i$. \\
            Update the dual variables, $\forall j \in \mathcal{N}_i$, $l \in \mathcal{E}_i$:
            \begin{align*}
                {\pmb{\lambda}_i^{(l)}}^+ &\leftarrow \pmb{\lambda}_i^{(l)} + \rho \mathbf{c}_i^{(l)}(\Bar{\mathbf{x}}, \Bar{\mathbf{w}} ) \\
                {\pmb{\mu}_i^{(j)}}^+ &\leftarrow \pmb{\mu}_i^{(j)} + \rho \mathbf{d}_i^{(j)}(\Bar{\mathbf{x}}, \Bar{\mathbf{w}} )
            \end{align*} \\
            Set cold start flag if any dual variable exceeds the predefined threshold
        \EndFor
        \Ensure Solution to local linearized problem, $\{ \hat{\mathbf{x}}^*[j] \}_{j \in \mathcal{N}'_i}$
    \end{algorithmic}
    \label{Algorithm:ADMM_Inner_Loop}
\end{algorithm}

\subsection{Analysis on Noise Effect}
\label{subsec:noise_effect}

The presence of noise was shown to be an issue with error reconstruction, occasionally resulting in divergence in initial numerical simulations. We use the penalty parameter $\rho$ to mitigate this undesirable behavior. Noisy measurements directly affect the constraints $\mathbf{c}( \cdot )$ and $\mathbf{d}( \cdot )$ in \eqref{eq:P3_lagrangian}. Thus, the penalty parameter $\rho$ can control the effect of noise. That is, reducing $\rho$ relaxes the constraints, thereby decreasing their effect on the objective function (\ref{eq:P3}). This effectively acts as a filter for the noisy measurements.
A similar approach to mitigating the effect of noise is theorized in \cite{khan2023recovery}. We validate this approach with extensive numerical simulations and mixed reality experiments in Section \ref{subsec:results_noise}.

\subsection{Analysis on Variations in Network Topology}
\label{subsec:varying_topology}

Through numerical simulations, we found that the algorithm could exhibit unstable behavior when the network topology is time-varying. Attributed to the warm start method, this would result in highly inaccurate reconstructed errors for a time-varying network topology. It was found that time dependent true and/or estimated states with the warm start method could result in large variations in the problem constraints and dual variables between iterations. Even simple configuration changes could cause the algorithm's dual variables to grow unbounded, causing divergence or poor reconstruction of the error. This is an expected consequence of constant constraint violations with ADMM \cite{boyd2011distributed}. 
To solve this issue, we propose adding a check to bypass the warm start technique when the dual variables grow beyond a heuristically determined threshold. When this check, referred to as ``cold start," is triggered, the dual variable values are reset to $\mathbf{0}$. This indicates the subsequent sub-problem is deemed sufficiently different enough to reset the problem parameters (i.e., the dual variables). The cold start method is validated in Section \ref{subsec:results_topology}. The dual variable resetting is handled on a per-robot basis, maintaining the distributed nature of the algorithm. 

\subsection{Integrity Monitor}

With the reconstructed error computed, we define a measure for the integrity of each robot $i \in \mathcal{V}$ to be 
\begin{equation}
    \label{eq:robot_integrity}
    \hat{{\chi}}_i = \norm{\hat{\mathbf{x}}[i] + \Bar{\mathbf{x}}[i]}_2.
\end{equation}
Note that at each iteration of the inner loop, robot $i$ communicates its integrity $\chi_i$ to a fault monitor. To account for noisy sensor measurements and transient behavior in the convergence of $\mathbf{x}$, we define a threshold value $\varepsilon$ for robot integrity. When $\hat{\chi}_i > \varepsilon$, we declare that a fault has been detected at robot $i$. The predefined threshold $\varepsilon$ is determined heuristically from the values of $\nu_\text{max}$ and $\omega_\text{max}$ to balance the detection accuracy and false detection. 

The integrity of robots $i \in \mathcal{V}$ are summed to measure the robot swarm's overall integrity:
\begin{equation}
    \label{eq:swarm_integrity}
    \hat{\chi} = \norm{\hat{\mathbf{x}} + \Bar{\mathbf{x}}}_{2,1}
\end{equation}
With the same threshold $\varepsilon$, we declare the integrity monitor has detected an attack on or fault in the MRS when $\hat{\chi} > \varepsilon$.

\section{Simulation Results}
\label{sec:sim}
\begin{figure}
    \centering
    \includegraphics[width=0.5\textwidth]{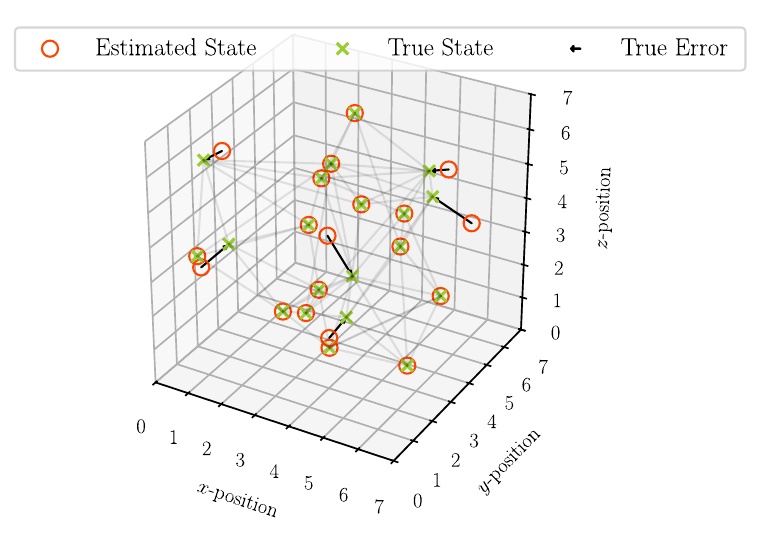}
    \caption{Example swarm configuration showing true and estimated robot states with true error vector}
    \label{fig:sim-config}
\end{figure}

\begin{figure*}[!h]
\centering
    \begin{subfigure}[t]{0.49\textwidth}
        \centering
        \includegraphics[width=\textwidth]{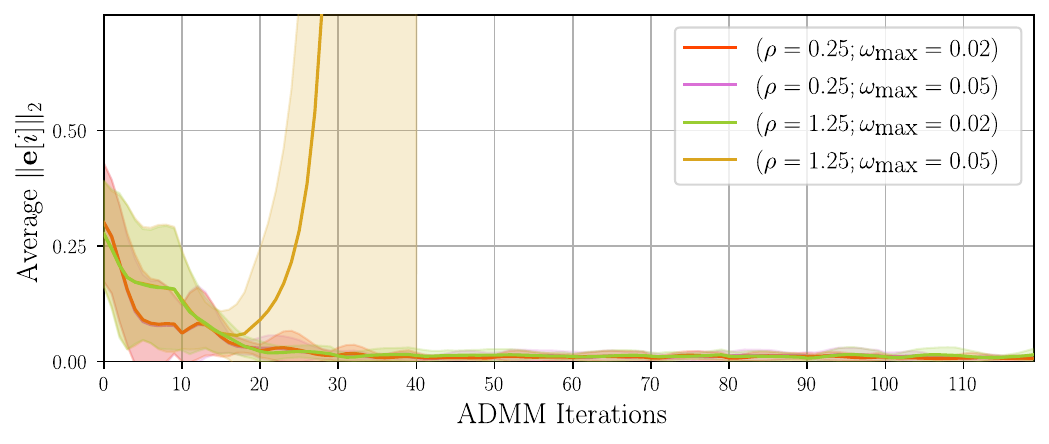}
        \caption{Comparing trials with varying $\rho$ and $\omega_\text{max}$ values}
        \label{fig:static-tests}
    \end{subfigure}
    \begin{subfigure}[t]{0.49\textwidth}
        \centering
        \includegraphics[width=\textwidth]{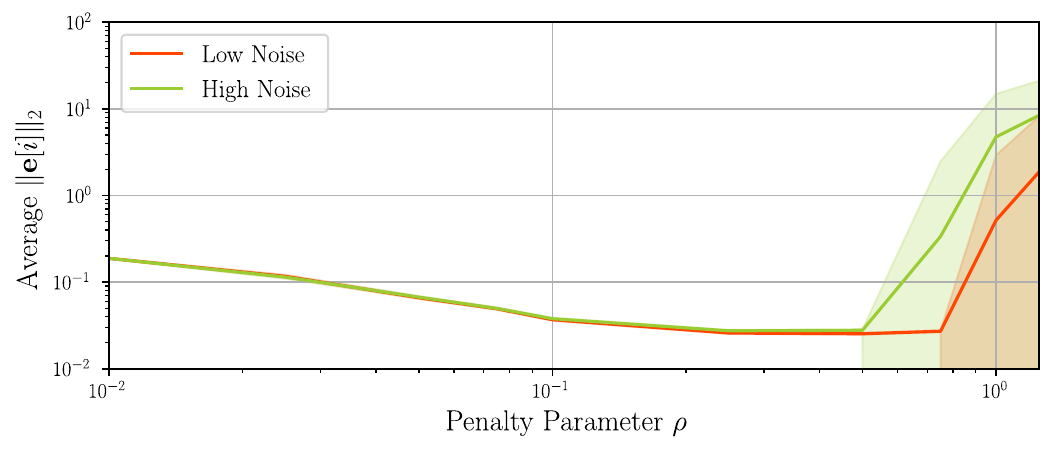}
        \caption{Monte Carlo Simulations with varying $\rho$ and $\omega_\text{max}$ values}
        \label{fig:static-monte}
    \end{subfigure}
\caption{Effect of noise on error reconstruction}
\label{fig:noise}
\end{figure*}

\begin{figure*}[!h]
\centering
    \begin{subfigure}[t]{0.49\textwidth}
        \centering
        \includegraphics[width=\textwidth]{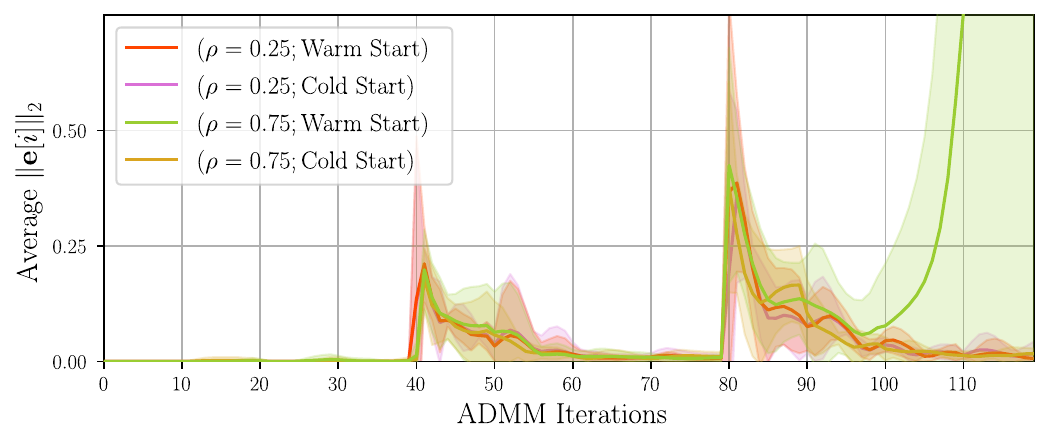}
        \caption{Comparing trials at varying $\rho$ values with warm vs. cold start}
        \label{fig:discrete-tests}
    \end{subfigure}
    \begin{subfigure}[t]{0.49\textwidth}
        \centering
        \includegraphics[width=\textwidth]{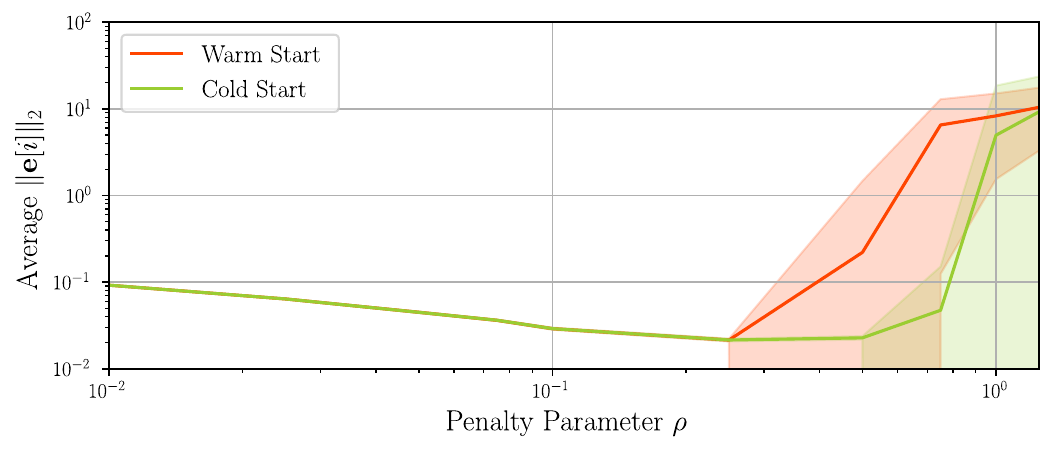}
        \caption{Monte Carlo Simulations at varying $\rho$ with warm vs. cold start}
        \label{fig:discrete-monte}
    \end{subfigure}
\caption{Effect of cold start on error reconstruction}
\label{fig:dynamic}
\end{figure*}

The proposed approach for multi-robot integrity monitoring is evaluated using numerical simulations and PX4-SiTL simulations on Gazebo. Through extensive Monte Carlo simulations, we validate the robustness of the proposed method to changes in the noise levels and time-varying network topologies. These issues are important as they occur during real-world deployment of an MRS but were not considered in the previous work \cite{khan2023collaborative}. For the numerical simulations, we consider a UAV swarm comprising $20$ UAVs, out of which $14$ are considered attack-free (healthy) and $6$ are considered attacked by an adversary using GNSS spoofing (see Fig. \ref{fig:sim-config}). To compare the results, we compute the difference between the reconstructed and true error vectors, denoted with $\mathbf{e}$, at each step in the simulation such that $ \mathbf{e} = \mathbf{x} - (\hat{\mathbf{x}} + \Bar{\mathbf{x}})$. A video of the overall  description and performance of our algorithm in simulations can be found at \href{https://tinyurl.com/mrs-integrity-monitor}{\texttt{https://tinyurl.com/mrs-integrity-monitor}}.

\subsection{Effect of Noise}
\label{subsec:results_noise}
In this section, we compare our algorithm's response to different constraint penalty parameters $\rho$ and inter-robot measurement noise $w_\text{max}$. To do so, we first conduct four simulations with the following parameter values and compare the error convergence properties: (i) $(\rho = 0.25, \omega_\text{max} = 0.02)$, (ii) $(\rho = 0.25, \omega_\text{max} = 0.05)$, (iii) $(\rho = 1.25, \omega_\text{max} = 0.02)$, and (iv) $(\rho = 1.25, \omega_\text{max} = 0.05)$. All tests are conducted with a maximum error in state estimation set at $\nu_\text{max}=0.02$. The root mean squared error (RMSE) $\| \mathbf{e}[i] \|_2$ is plotted in Fig. \ref{fig:discrete-tests}, with the darker lines indicating the average RMSE over the agents and the lighter shaded region representing $1$ standard deviation bounds. We see that while parameter configurations (i) - (iii) exhibit similar convergence, simulation (iv), with high noise and $\rho$, failed to converge with the reconstructed error approaching infinity. This hints at a relationship between the noise level, parameter $\rho$, and convergence. 
For further investigation, we conducted Monte Carlo simulations with $\omega_\text{max} = 0.02$ and $\omega_\text{max} = 0.05$ along various values of $\rho$. Each simulation configuration was conducted for $100$ trials. The RMSE was computed from each trial and averaged over the agents and trials. The results, plotted in Fig. \ref{fig:static-monte}, show convergence to the true error is diminished when higher values of $\rho$ are selected with high noise levels. Thus, in a noisy system, we must seek to balance a high convergence rate (achieved with higher $\rho$ values) with low divergence chances (achieved with lower $\rho$ values).

\subsection{Effect of Variations in Network Topology}
\label{subsec:results_topology}

We consider a time-varying network topology induced by a $2$-phase cyberattack, with only $3$ robots being affected in each phase. Noise levels are set in all trials at $(\nu_\text{max} = 0.02, \omega_\text{max} = 0.02)$. The proposed algorithm's performance is compared under warm and cold start with $\rho = 0.25$ and $\rho=0.75$. The RMSE $\| \mathbf{e}[i]\|_2$ is computed for each simulation and plotted in Fig. \ref{fig:discrete-tests} in a similar fashion as Fig. \ref{fig:static-tests}. We see that $3$ simulation configurations exhibit similar error reconstruction properties, while the simulation with $\rho = 0.75$ under warm start diverges after the second attack phase is engaged. The robustness of cold start under the changing network topology was further investigated with Monte Carlo simulations at various values of $\rho$ under warm and cold start, with $100$ trials for each simulation configuration. The results are plotted in Fig. \ref{fig:discrete-monte} and show that there isn't a significant disparity between warm and cold start for low $\rho$. This increases for large $\rho$, with cold start significantly outperforming warm start, demonstrating its increased robustness against time-varying network topology.

\section{Experimental Demonstration}
\label{sec:exp}
\subsection{Sensor Emulation for UWB Range Measurements}
\begin{figure}[h!] 
\centering
  \includegraphics[width=0.45\textwidth]{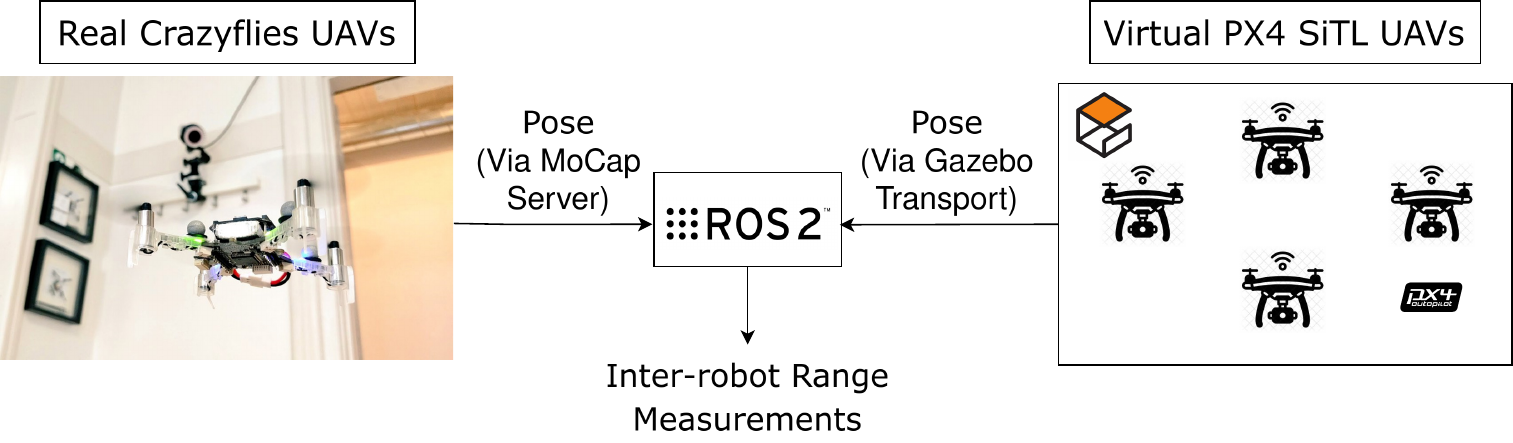}
  \caption{Pose information is gathered from MoCap and Gazebo to emulate the inter-robot range measurements.}
  \label{fig:range_emulation}
\end{figure}

We leverage our Mixed Reality sensor emulation framework, i.e., Mixed-Sense \cite{pant2024mixed}, to generate inter-robot range measurements as shown in Fig. \ref{fig:range_emulation}. In the Mixed-Sense framework, we create digital twins of real robots, tracked using motion capture cameras, inside Gazebo. Additionally, we spawn virtual simulated robots along with real robots in real-time. This allows us to create an augmented reality for the real robots, where these robots, although moving inside the controlled motion capture volume, get the sensory information from a simulated 3D environment. The sensor measurements from the simulator are transported to the real robots in real-time, allowing real robots to operate together with virtual robots inside the mixed reality environment. This feature enables us to validate the (i) scalability by spawning a large number of virtual robots alongside real robots, and (ii) interoperability by choosing various types of SiTL instances with the real robots. 
To generate the inter-robot range measurements, the pose of real robots tracked by the motion capture system is transported to Gazebo via ROS2. The pose of the virtual robots is retrieved from Gazebo's transport. The block diagram in Fig. \ref{fig:range_emulation} shows the overall architecture employed to generate emulated range measurements. A ROS node utilizes the poses of real and virtual robots to generate the inter-robot range measurement and publish it as a ROS topic.

\subsection{Mixed Reality Experiments}
In our mixed reality setup, we use four real Crazyflies UAVs and spawn three PX4-SiTL instances. The real UAVs are tracked using the Qualisys Motion Capture system, which consists of $6$ Oqus 7+ cameras. We consider the scenario where an adversary spoofs one real and one PX4-SiTL drone. The position of these drones drifts, affecting the safety of the overall swarm operation. The simulator hosting Gazebo is launched using a Dell Precision-3581 workstation laptop with a 13th-Gen Intel Core i7 processor and an Nvidia RTX A1000 GPU, where inter-robot range measurements are also emulated. We use Crazyflie 2.0 swarms as real UAVs for our demonstration. We use the open-source PX4 autopilot to instantiate virtual SiTL UAVs in Gazebo.

\begin{figure*}[!h] 
\centering
  \includegraphics[width = \textwidth]{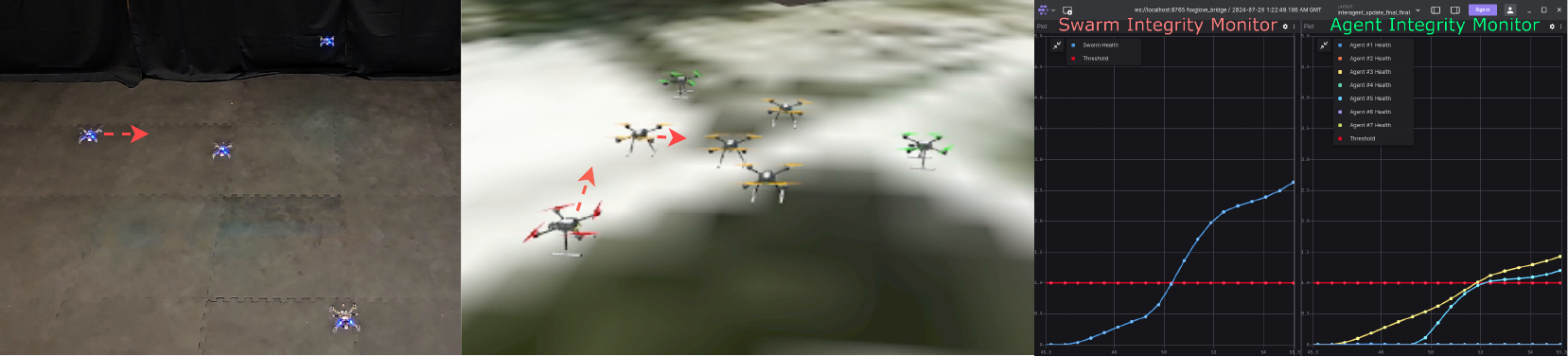}
  \caption{Mixed reality experiments showcasing the effectiveness of our proposed framework. \textbf{Left}: Real Crazyflie swarm tracked by motion capture system. \textbf{Center}: Real + Virtual UAVs in Gazebo performing a formation flight mission. \textbf{Right}: Integrity monitors showing real-time integrity measure for the overall swarm as well as individual agents.}
  \label{fig:mixed_reality}
\end{figure*}

Under nominal operations, the swarm, comprising real and virtual UAVs, is commanded to first takeoff, hover, and transition into a static formation flight shown in Fig. \ref{fig:mixed_reality}. For the experimental demonstration, the integrity monitoring algorithm is implemented centrally, where all the information is aggregated at the ground station, i.e., the Dell Precision-3581 workstation laptop. The adversary then spoofs the GNSS sensor of two UAVs, one real and one virtual. We programmatically add a constant offset in the GNSS coordinates to alter the position measurements of both drones shown in Fig. \ref{fig:mixed_reality}. The integrity monitor for each vehicle detects an attack when the integrity measure crosses the threshold. The video found at \href{https://tinyurl.com/mrs-integrity-monitor}{\texttt{https://tinyurl.com/mrs-integrity-monitor}} also validates the algorithm using mixed reality experiments.

\section{Conclusion}
\label{sec:conclusion}
In this paper, we proposed a multi-robot integrity monitoring algorithm that utilizes locally sensed inter-robot range measurements to detect and identify the presence of rogue or faulty robots in the system. We validated the robustness of the proposed algorithm to noise and time-varying network topologies through extensive numerical Monte Carlo simulations. Finally, we demonstrated the effectiveness of our algorithm through mixed-reality experiments on a heterogeneous robot swarm. 
In the future, we will demonstrate a distributed implementation of our algorithm in mixed-reality, allowing each agent to implement it independently using its locally observed information. We also plan to investigate the use of other inter-robot measurements, such as bearing and subtended angle, and integrate them with our algorithm to make it more robust against cyberattacks or faults. 

\bibliographystyle{ieeetr}
\bibliography{IEEEabrv,bibs} 

\end{document}